\pdfoutput=1

\documentclass[10pt,journal,compsoc]{IEEEtran}
\ifCLASSOPTIONcompsoc
  \usepackage[nocompress]{cite}
\else
  \usepackage{cite}
\fi
\ifCLASSINFOpdf
\else
\fi
\usepackage{subfig}
\usepackage{graphicx}
\usepackage{comment}
\usepackage{amsfonts}
\usepackage{amsmath}
\usepackage{algorithm}
\usepackage{algorithmic}
\usepackage{bm}
\DeclareMathOperator*{\argmax}{arg\,max}

\usepackage{multirow}
\begin{document}
%
\title{Plug-and-Play Anomaly Detection with \\ Expectation Maximization Filtering}
%
%
%
%

\author{Muhammad Umar Karim Khan,
        Mishal Fatima,
        and~Chong-Min~Kyung,~\IEEEmembership{Fellow,~IEEE}
\IEEEcompsocitemizethanks{\IEEEcompsocthanksitem Muhammad Umar Karim Khan is with the Center of Integrated Smart Sensors, KAIST, 34141.\protect \\
E-mail: umar@kaist.ac.kr
\IEEEcompsocthanksitem Mishal Fatima and Chong-Min Kyung are with the Department of Electrical Engineering, KAIST.\protect \\
E-mail: {mishal, kyung}@ee.kaist.ac.kr}
\thanks{This work was supported by ICT grant CISS-2013073718}}

%
%

\markboth{XXXXXXXXXXXX,~Vol.~XX, No.~X, August~XXXX}%
{XXXX \MakeLowercase{\textit{et al.}}: Plug-and-Play Anomaly Detection with Expectations Maximization Filtering}
%



\IEEEtitleabstractindextext{%
\begin{abstract}
Anomaly detection in crowds enables early rescue response. A plug-and-play smart camera for crowd surveillance has numerous constraints different from typical anomaly detection: the training data cannot be used iteratively; there are no training labels; and training and classification needs to be performed simultaneously. We tackle all these constraints with our approach in this paper. We propose a Core Anomaly-Detection (CAD) neural network which learns the motion behavior of objects in the scene with an unsupervised method. On average over standard datasets, CAD with a single epoch of training shows a percentage increase in Area Under the Curve (AUC) of 4.66\% and 4.9\% compared to the best results with convolutional autoencoders and convolutional LSTM-based methods, respectively. With a single epoch of training, our method improves the AUC by 8.03\% compared to the convolutional LSTM-based approach. We also propose an Expectation Maximization filter which chooses samples for training the core anomaly-detection network. The overall framework improves the AUC compared to future frame prediction-based approach by 24.87\% when crowd anomaly detection is performed on a video stream. We believe our work is the first step towards using deep learning methods with autonomous plug-and-play smart cameras for crowd anomaly detection.
\end{abstract}

}

\maketitle

\IEEEdisplaynontitleabstractindextext

%
\IEEEpeerreviewmaketitle

\IEEEraisesectionheading{\section{Introduction}\label{sec:introduction}}

%
%
%
%
\IEEEPARstart{S}{urveillance} plays a key role in maintaining healthy societies. Authorities use it to detect accidents and take necessary precautionary measures. Surveillance is necessary for a justice system to operate and to minimize criminal incidents. With the increase in population and technology, emphasis on automated surveillance has been increasing rapidly. Vision-based approaches have so far attracted researchers the most as videos contain more information compared to other sources of information to comprehend an activity. Thus, automated vision-based surveillance can play an important role in early response to accidents and criminal activities.

Accidents in crowds can be catastrophic. The large density of people can lead to a large number of casualties even if the accident is limited to a small area. Furthermore, the probability of accidents happening in crowds is much higher. Automatic detection of accidents or events in crowds provides a critical opportunity to avoid or minimize casualties.

Anomaly detection is a viable approach for crowd surveillance. In anomaly detection, a system is trained to learn the normal behavior of the input. An anomaly is declared if the observed behavior is different from the learned normal behavior. The normal behavior of crowds is generally quite redundant, whereas an accident can happen from a variety of sources. Even training a neural network with a large number of accident cases does not guarantee accurate accident detection, as accidents may not occur from sources seen during training. Furthermore, more videos containing normal crowd behavior are available than videos of accidents as accidents occur rarely. This imbalance of data can hamper optimal training of a classification network. These issues make anomaly detection a reasonable choice for crowd surveillance as it requires videos containing only normal behavior for training, and, theoretically, can detect any kind of accident.

Numerous approaches have been presented in the past for anomaly detection in crowds. These approaches have separate training and testing stages. Normal behavior is learned at the training stage by only using normal examples. Classical approaches use handcrafted features and machine learning to model normal crowd behavior. With the advent of deep learning, researchers have shifted towards deep neural networks for crowd anomaly detection and obtained significantly better results.

Plug-and-play operation requires anomaly detection to be operable in any location where it is installed. A typical approach in this scenario is to train a neural network with a large amount of training data obtained from many different scenes. Along with the difficulty of gathering training data, using a single neural network for anomaly detection with different scenes does not produce good results. In fact, there have been no reports of research where the same neural network has been used across multiple views for crowd anomaly detection. Thus, using training data from the scene where the anomaly detection system is deployed is highly desired.

A plug-and-play anomaly detection system needs to learn the normal behavior of the scene where it is deployed. This results in a video stream of data to be used in real-time for training the anomaly detection system. For the sake of efficiency and synchronization, a training sample once used cannot be used again. This approach is called continual learning \cite{parisi2019continual}, and is different from typical crowd anomaly detection where the training data is stored and repeatedly used for training the neural network.

Another constraint on a plug-and-play anomaly detection system is to learn and classify at the same time. There is no guarantee that the input contains only normal data to be used for training. To train the neural network, it is necessary to divide the input samples into normal and anomalous, and use the normal samples for training. However, the same neural network is used for classification. Simply put, the training data depends on the anomaly detection performance of the neural network and the anomaly detection performance depends on the training data. This makes it a typical chicken-and-egg problem.

In this paper, we make an effort towards a plug-and-play crowd anomaly detection system that can deal with the above constraints. Our key contributions are as follows.
\begin{itemize}
\item We propose a core anomaly-detection system, which learns the normal motion behavior of objects in the scene. We use an unsupervised approach to learn the motion behavior of the objects. The core anomaly-detection system shows competitive performance even with a single epoch of training. 

\item We propose a filter based on the Expectation-Maximization (EM) algorithm to be used in conjunction with the core anomaly-detection system. The proposed filter filters-out anomalous samples such that only normal samples are used for training the core anomaly-detection system.
\end{itemize}

\begin{figure*} [!bt] 
\centering
\includegraphics[width=0.9\textwidth]{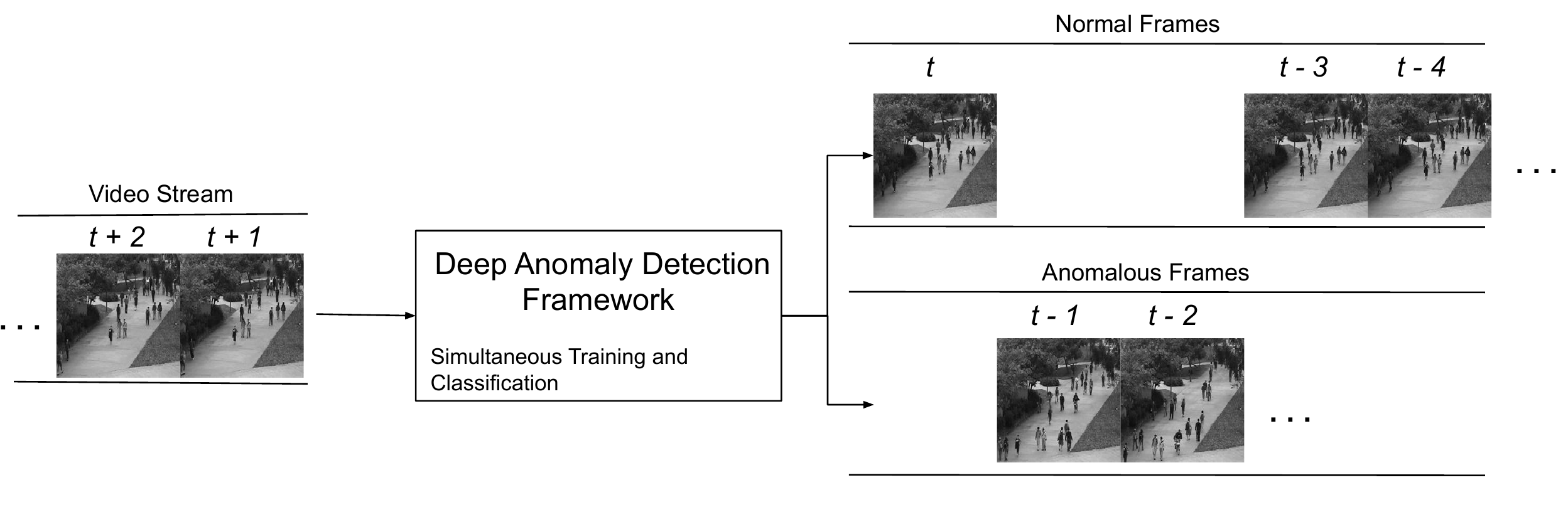}
\caption{Crowd anomaly detection on an input video stream. The video stream is not stored for later use. The anomaly detection system needs to learn and classify simultaneously.}
\label{fig:problem_fig}
\end{figure*}

\noindent In the past, online approaches for crowd anomaly detection have been proposed. However, these approaches are generally based on classical algorithms and do not leverage the power of deep learning. Also, these approaches generally have a shorter memory compared to a deep neural network. In other words, these approaches use salience in a limited spatial and temporal neighborhood of pixels. Our approach has a longer memory as it uses a deep neural network to learn the normal behavior in a given scene. The operation of the proposed method is shown in Fig. \ref{fig:problem_fig}. To our knowledge, this is the first effort towards online crowd-anomaly detection with deep learning.

The rest of the paper is structured as follows. Research related to our work is discussed in Section 2. In Section 3, we describe the core anomaly-detection system. The EM filter is discussed in Section 4. Experimental results and discussion are given in Section 5, and Section 6 concludes the paper.

\section{Related Works} \label{sec:rel_work}
\noindent Anomaly detection has been used with both one-class and multi-class problems. Other terms have also been used for anomaly detection in literature, which include outlier detection, out-of-distribution detection and novelty detection. In \cite{hendrycks2016baseline}, the authors show that deep neural networks trained for a multi-class problem can be used for detecting out-of-distribution samples. They propose an abnormality detection module, which can be used with multi-class deep neural networks. In \cite{lee2018simple}, the layer-wise features of a classification neural network are assumed to have a Gaussian distribution. The authors estimate the parameters of the Gaussian distribution for the features at every layer and use the Mahalanobis distance to detect out-of-distribution samples. Authors in \cite{xu2010robust} propose a convex optimization approach to detects outliers. They consider the similarity between the low-dimensional approximation of the inputs using PCA. In \cite{xia2015learning}, the reconstruction error of an autoencoder is used to detect outlier images. The authors enhance the performance of the autoencoder by training it with examples that were determined to be normal in the past epochs. In \cite{you2017provable}, authors consider sparse representations of data samples to detect outliers. The authors in \cite{sabokrou2018adversarially} input the images reconstructed by a U-net \cite{isola2017image} to the discriminator of a GAN \cite{goodfellow2014generative} to detect outliers.

Classical approaches have used hand-crafted features for crowd anomaly detection. In \cite{mahadevan2010anomaly}, authors propose Mixture of Dynamic Textures (MDT) for crowd anomaly detection, which considers both the appearance and motion behavior of objects. Their work is further enhanced in \cite{li2014anomaly}. Another feature called HOFME \cite{colque2017histograms} (which is an extension of HOFM \cite{colque2015histograms}) considers the histogram of optical flow's direction, optical flow's magnitude and entropy for crowd anomaly detection. MPPCA \cite{kim2009observe} considers optical flow in grids of local regions based on space and time as nodes of a Hidden Markov Random Field (MRF). This approach detects both local and global anomalies using the MRF.  In \cite{mousavi2015abnormality} and \cite{mousavi2015analyzing}, the authors estimate statistics of tracklets in fixed cuboids to estimate normal crowd behavior.  \cite{mehran2009abnormal} uses the social force model to model the behavior of crowds. Anomaly is declared if the observed behavior is different from the presumed model. \cite{masoudirad2017anomaly} uses a sparse representation of video frames with a two-part dictionary to achieve very high speed of anomaly detection. \cite{lin2015anomaly} uses one-class SVM for crowd anomaly detection. \cite{leyva2017abnormal} proposes binary features for fast speed and uses Gaussian Mixture Models (GMM) for fast anomaly detection. The authors further enhance their work in \cite{leyva2017video}. In \cite{li2015spatio}, features based on histogram of gradients of cuboids are compared against both spatially and temporally neighboring cuboids to detect anomalies. \cite{chaker2017social} proposes using local social networks with cuboids from the video and a global social network, which is updated based on the local social networks. Authors in \cite{rojas2016abnormal} model the optical flow of moving objects in windows of the video frame using a GMM. An anomaly is declared by considering the Mahalanobis distance of a given sample from every constituent of the mixture model. In \cite{Lu2019}, authors propose a fast sparse-dictionary approach for crowd anomaly detection. They also propose a mini-batch scheme for reduced memory consumption while learning a dictionary. Authors in \cite{del2016discriminative} propose an approach where the anomaly detection performance is independent of the order of occurrence of the anomaly. They train a logistic regressor on sliding windows of multiple features to detect anomalies.

Approaches based on deep learning have significantly improved the overall accuracy of crowd anomaly detection. These approaches are generally unsupervised, i.e., the anomalous samples are not observed during training. In Appearance and Motion DeepNet (AMDN) \cite{xu2017detecting}, authors use autoencoders to extract appearance and motion features separately. These features are combined and classified using a single-class SVM. A Restricted Boltzmann's Machine (RBM) has been used in \cite{vu2017energy} to model normal crowd behavior. \cite{sun2017abnormal} proposes an end-to-end deep learning network called Deep One Class (DOC) to detect anomalies. Authors in \cite{hasan2016learning} use convolutional autoencoders (Conv. AE) for anomaly detection. They consider both the actual video as well as handcrafted features as input to the autoencoders and use the reconstruction error to determine an anomaly. \cite{hu2016video} uses a deep incremental slow feature analysis network to extract features for anomaly detection and classify a video as anomalous or not. \cite{luo2017remembering} uses a convolutional LSTM with an autoencoder (ConvLSTM-AE) to reconstruct the current and the previous frames of the normal samples. Reconstruction error is expected to be high with the anomalous samples, thus, allowing detection of anomalies. \cite{tudor2017unmasking} uses the unmasking approach applied to texts in the past \cite{koppel2007measuring} for crowd anomaly detection. Pre-trained networks have been used in \cite{SABOKROU2018} for feature extraction. A sparse autoencoder has been used in \cite{sabokrou2016video} for anomaly detection. Authors in \cite{chong2017abnormal} propose a unified autoencoder learning both the spatial and temporal features through different layers for crowd anomaly detection. \cite{luo2017revisit} proposes temporally-coherent sparse coding and uses it with a recurrent neural network for anomaly detection. In \cite{ravanbakhsh2016plug}, authors propose using a binary fully-convolutional neural network to extract binary maps from a video. Optical flow is used with these maps to detect an anomaly. Authors in \cite{liu2018future} predict future frames using a network similar to Pix2Pix \cite{isola2017image}. They input a number of previous frames to the U-net and compare the predicted frame with the actual frame for anomaly detection. In \cite{ravanbakhsh2019training}, authors perform crowd anomaly detection using cyclic GANs. One half cycle predicts the future frame while the other half cycle predicts the optical flow using the future frame. Both the predicted future frame and optical flow are compared against their original counterparts, and an anomaly is declared if the difference is too large. A similar approach is adopted by the authors in \cite{vu2019robust}; however, their approach is not cyclic. In \cite{ionescu2019object}, the authors first extract objects from a video frame, and train an autoencoder with the appearance and optical flow of the objects. Their approach does not consider the context of the objects. Although results achieved by deep-learning methods for crowd anomaly detection are quite accurate, these methods do not show good performance under the constraints discussed in Section 1. 

Continual learning provides deep neural networks with the ability to learn over their lifetime. In continual learning, the system should be capable of learning from a stream of data without having direct access to past data. The same constraint is posed by a plug-and-play crowd anomaly detection system. One of the challenges faced by deep neural networks with continual learning is catastrophic forgetting, when the neural network is strongly influenced by recent data and forgets the past data. Numerous techniques have been proposed for continual learning, which include regularization over time, extendable neural networks, dual memory systems, curriculum learning, transfer learning and cross-modal learning. Further details of continual learning schemes can be found in the survey \cite{parisi2019continual}. However, the capability of continual learning typically comes with additional computational complexity.

Some researchers have focused on the problem of online crowd anomaly detection, which requires anomaly detection on-the-fly without prior training. The authors in \cite{sikdar2019adaptive} propose a training-less system, which compares the motion characteristics of pedestrians across consecutive frames using the earth mover distance. An abrupt change in motion characteristics indicates an anomaly. Similarly, authors in \cite{yuan2014online} perform online anomaly detection by proposing a structural context descriptor. They also consider the variation in the descriptor by comparing consecutive frames using the earth mover distance. \cite{dutta2015online} performs an online update to the dictionary of sparse coefficients. Authors in \cite{feng2010online} propose using a $8 \times 8$ self-organizing map for online crowd anomaly detection. In \cite{sun2017online}, the authors propose using growing neural-gas network \cite{fritzke1995growing} for this problem. They first train a neural network by using the training data. Afterwards, they further improve the neural network at testing stage; thus, the method is termed online. The maximum number of neurons is limited to 300. \cite{pennisi2016online} considers the Shannon entropy of video features. They claim that an irregular event shows higher entropy compared to regular events. In \cite{javan2013online}, histograms of oriented gradients (spatial and temporal) are clustered in an online manner for online crowd anomaly detection. In \cite{bera2016realtime}, motion trajectories of individuals are compared against other pedestrians in a scene to detect an anomaly. 

From the literature, it is seen that most of the methods used for crowd anomaly detection cannot work well for plug-and-play operation. General purpose continual learning schemes may allow plug-and-play operation but only at the additional computational complexity. Most of the online anomaly detection schemes proposed in literature are based on handcrafted features and detect anomalies using the salience of features in a local spatial and temporal neighborhood. These methods may fail if anomaly occurs for a relatively longer duration or most of the objects show anomalous behavior at the same time. On the other hand, deep neural networks have a much longer memory of events. Our work proposes a continual learning approach for learning the normal behavior using a deep neural network while classifying anomalous events at the same time.

\section{Core Anomaly Detector} \label{sec:core_anomaly_detector}
\noindent In this section, we describe the anomaly detector that is to be used as a component in the overall plug-and-play system. The Core Anomaly Detector (CAD) is expected to work under typical evaluation conditions. In detail, CAD can be trained with normal data at train time and used to detect anomalies at test time. However, unlike most of the previous anomaly detection methods, CAD is expected to work with a stream of training data. It is not allowed to repeatedly use past samples for training and should continually learn the normal behavior in a given scene. This is equivalent to saying that CAD should show reasonable performance with training over a single epoch of the training samples.

Both motion and appearance are important cues to determine an object's behavior. A change in velocity or appearance of objects in a crowd can be used to declare an anomaly. In the past, researchers have explicitly combined both appearance and motion information in neural networks to detect anomalous crowd behavior \cite{xu2017detecting}, \cite{hasan2016learning}. This adds to computational complexity as two different pipelines are required. In this work, we use a single pipeline which uses motion to predict the future frame. By using this approach, we get two distinct advantages. First, the neural network learns both the motion and appearance with a single pipeline. Second, the neural network is trained to generate the optical flow, which is generally sparse and redundant. Smaller parameter updates are required by the neural network to learn to generate the optical flow compared to the parameter updates required in generating the actual frames. This is due to less variation in the motion behavior within each category of objects despite the large variation in appearance. For example, cars of different shapes on the same road have very similar optical flow. As a result, the variation in weight parameters over time is much smaller leading to less catastrophic forgetting \cite{kemker2018measuring}. Since the motion of the objects is not as diverse as their appearance, the neural network learns quickly as there is little information to learn.

\begin{figure}[!bt]
\includegraphics[width=\columnwidth]{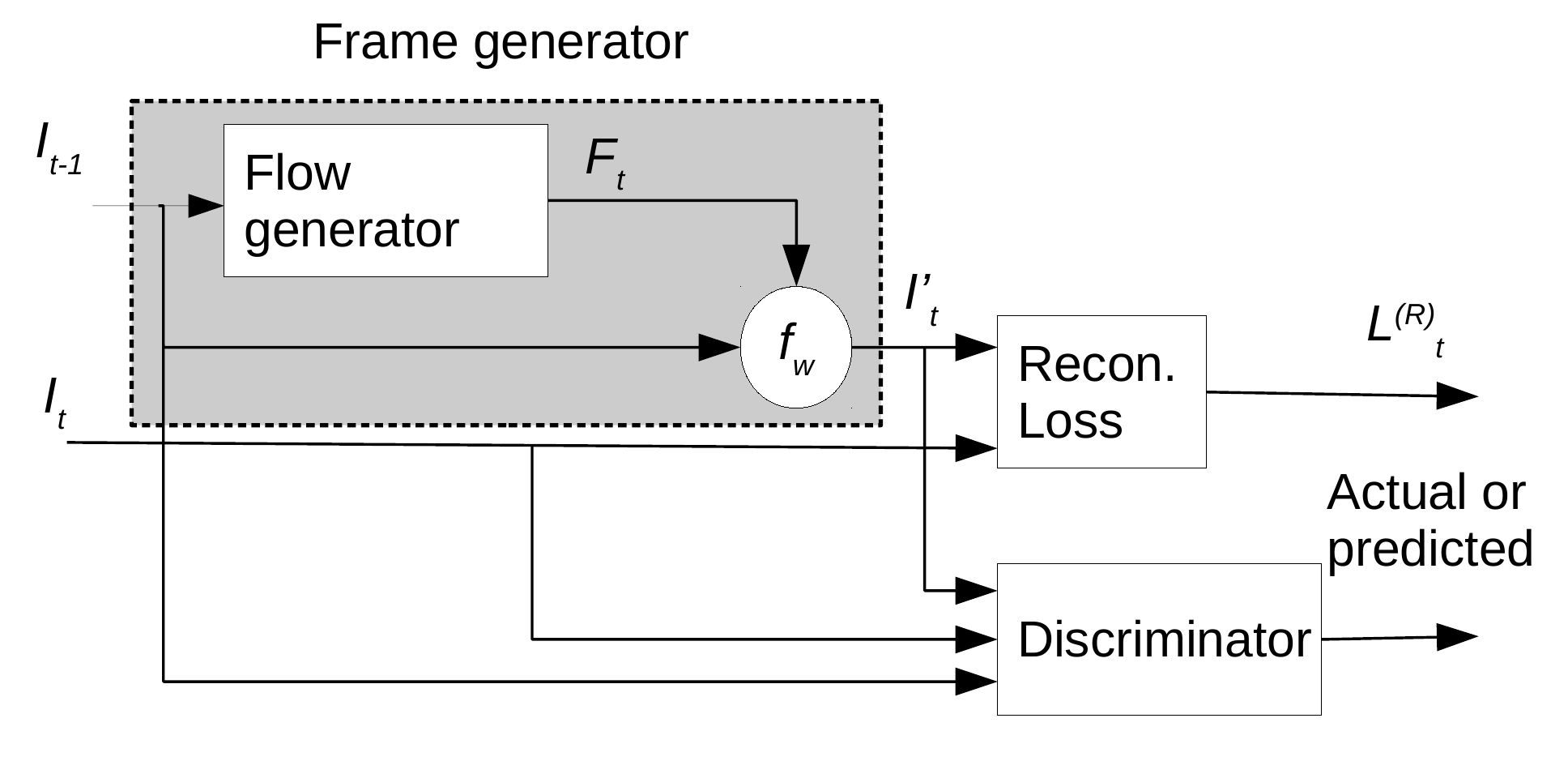}
\caption{The Core Anomaly Detector.}
\label{fig:core_ano_det}
\end{figure}

The proposed system for learning the motion behavior is shown in Fig. \ref{fig:core_ano_det}. The input to the system are two consecutive frames, $I_{t-1}$ and $I_t$. The frame generator predicts the current frame $I'_t$ with $I_{t-1}$ as the input and compares $I'_t$ against $I_t$. Key component of the frame generator is the flow generator, which is used to generate the optical flow $F_t$ between $I_{t-1}$ and $I_t$. It should be noted that the only input to the flow generator is the previous frame, $I_{t-1}$, i.e., the flow generator generates optical flow using a single frame. This is only possible if the flow generator remembers the optical flow associated with objects in the normal samples. In summary, the neural network learns the motion behavior of objects in a scene if it is required to predict optical flow using a single frame. The current frame is reconstructed from the previous frame and the optical flow using the warping function \cite{jaderberg2015spatial} as

\begin{equation} \label{eq:warp_function}
I'_t = f_w(I_{t-1}; F_t),
\end{equation}

\noindent where $f_w$ is the warping function. Note that the warping function is fully differentiable.

The parameters of the neural network are adjusted to minimize the reconstruction error between the predicted current frame $I'_t$ and the actual current frame $I_t$. Our first loss function is the difference of the image intensities called the intensity loss defined as

\begin{equation}\label{eq:int_loss}
L_t^{(i)} = \sum_j \left(I_t(j) - I'_t(j)\right)^2,
\end{equation}

\noindent where $j$ is the pixel index. We also minimize the loss between the first order image-gradients of the predicted and actual frames. The gradient loss is defined as 

\begin{equation}\label{eq:grad_loss}
L_t^{(g)} = \sum_j \left((\nabla I_t)_x(j) - (\nabla I'_t)_x(j)\right)^2,
\end{equation}

\noindent where $(\nabla I_t)_x$ is obtained after applying the filter $[1, 0, -1]$ to the $I_t$. The reconstruction loss is defined as the sum of the intensity loss and gradient loss, i.e., 

\begin{equation}\label{eq:recon_loss}
L^{(R)}_t = L_t^{(i)} + L_t^{(g)}.
\end{equation}

Using the ideas presented in \cite{isola2017image}, we use the GAN principle \cite{goodfellow2014generative} to enhance the performance of the frame generator. The discriminator is trained to differentiate between the predicted $I'_t$ and actual $I_t$ by minimizing the following loss function.

\begin{equation}\label{eq:disc_loss}
\begin{aligned}
L_t^{(D)}(G, D, I_{t-1}, I_{t}) = -\mathbb{E}_{I_t \sim p(I_t)}[\log D(I_t)] \\
 -\mathbb{E}_{I_{t-1}\sim p(I_{t-1})}[log(1-D(I'_t)],
\end{aligned}
\end{equation}

\noindent where $G$ and $D$ represent the frame generator and the discriminator, respectively, and

\begin{equation}\label{eq:gan_equalities}
I'_t = G(I_{t-1}).
\end{equation}

\noindent By minimizing the first term on the RHS of (\ref{eq:disc_loss}), the discriminator will return a value close to 1 if the actual current frame $I_t$ is input to the discriminator. On the contrary, by minimizing the second term on the RHS of (\ref{eq:disc_loss}) the discriminator will return a value close to 0 if the predicted current frame $I'_t$ is input to the discriminator. Similarly, the following loss function is used to train the generator. 

\begin{equation} \label{eq:gen_loss}
L_t^{(G)}(G, D, I_{t-1}) = -\mathbb{E}_{I_{t-1}\sim p(I_{t-1})} [\log D(I'_t)].
\end{equation}

\noindent The RHS of (\ref{eq:gen_loss}) will be reduced if the predicted current frame $I'_t$ inclines the discriminator to produce an output of 1. In other words, the frame generator is trained to produce current frames that fool the discriminator. Training is performed repeatedly in two steps. In the first step the parameters of the discriminator are updated to minimize the loss function in (\ref{eq:disc_loss}), while in the second step the sum of the reconstruction loss $L_t^{(R)}$ and $L_t^{(G)}$ from (\ref{eq:gen_loss}), i.e., 

\begin{equation} \label{eq:ov_gen_loss}
L_{t}^{(O)} = L_t^{(R)} + \lambda L_t^{(G)},
\end{equation}

\noindent is minimized. Here $\lambda$ is the weighting factor. We use the reconstruction loss $L_t^{(R)}$ in (\ref{eq:recon_loss}) at inference to detect an anomaly.

\begin{figure*}[!bt]
\centering
\includegraphics[width=0.9\textwidth]{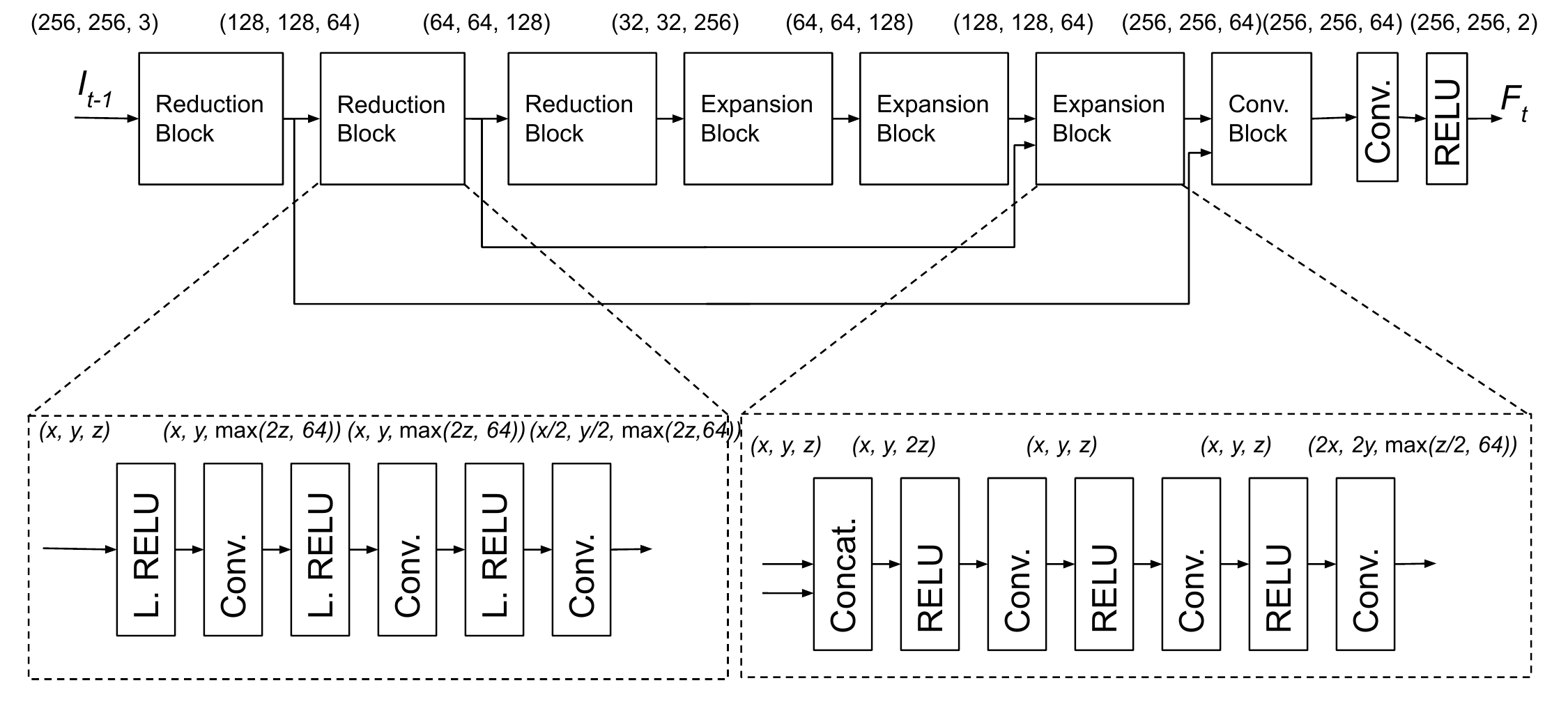}
\caption{Flow generator of CAD. Conv. block is the same as reduction block except it maintains the spatial resolution and number of features of the input at the output. All convolutional filters are $5\times 5$.}
\label{fig:gen_layers}
\end{figure*}

\begin{figure}[!bt] 
\centering
\includegraphics[width=1.0\columnwidth]{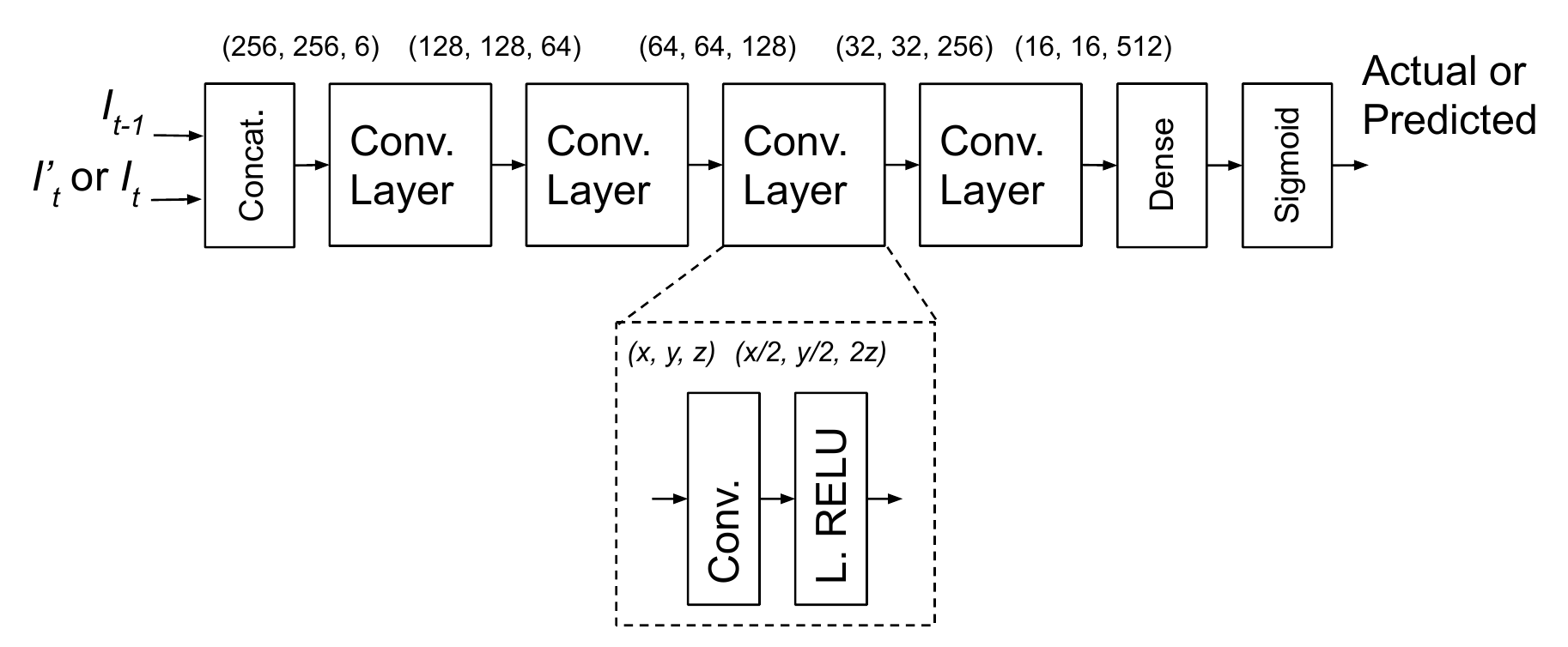}
\caption{Discriminator of CAD. All convolutional filters are $5 \times 5$.}
\label{fig:disc_layers}
\end{figure}

Detailed operations of the flow generator and discriminator are shown in Fig. \ref{fig:gen_layers} and \ref{fig:disc_layers}, respectively. Training the proposed approach is completely unsupervised, i.e., the ground truth optical flow is not required for training the overall system. The network used in \cite{liu2018future} also predicts frames to detect anomalies. However, it generates the future frame by considering the previous five frames. Also, the authors do not use optical flow to reconstruct the frame as in our approach. Rather, they try to reduce the perceptual loss \cite{johnson2016perceptual} between the generated frames and Flow-Net \cite{dosovitskiy2015flownet}, as well as the frame reconstruction loss. CAD has a single pipeline whereas their network has two pipelines, one for extracting Flow-Net features and the other for predicting the future frame.

Unsupervised learning of optical flow has been suggested earlier \cite{ren2017unsupervised} and the warping function has also been used for disparity estimation in stereo images \cite{godard2017unsupervised}. However, the critical difference between our approach and \cite{ren2017unsupervised} is that \cite{ren2017unsupervised} inputs two consecutive frames to an optical flow generator whereas we input a single frame to the flow generator. Thus, the neural network in \cite{ren2017unsupervised} learns the difference between two consecutive frames, whereas CAD memorizes the motion associated with appearance in a given frame. There are other differences between \cite{ren2017unsupervised} and CAD. The GAN principle was not used for training in their method. Also, the neural architecture of CAD is different from that proposed in \cite{ren2017unsupervised}. Furthermore, we do not use the multi-scale loss as it does not improve the performance of CAD.

\section{Expectation Maximization Filtering} \label{sec:em_filtering}
\noindent Despite its advantages, CAD presented in the previous section cannot deal with all the challenges described in the introductory section. CAD can only perform anomaly detection if it is trained with normal samples. However, unlike conventional methods discussed in literature, training and classification needs to be performed simultaneously in a plug-and-play anomaly detection system. Both these tasks are inter-dependent. We need a well-trained anomaly detector to classify samples into normal and anomalous. On the other hand, we need good classification to pass normal samples to the anomaly detector for training. This section proposes a solution to this problem.

The EM algorithm consisting of the E-step and the M-step has been extensively used in the past to estimate the parameters of a model when the model depends on latent variables. Let $X = [x_1, x_2, ..., x_n]$ be the set of the input samples including both normal and anomalous, and $Y=[y_1, y_2, ..., y_n]$ be the set of corresponding latent variables identifying an anomaly. More specifically, $y_i \in \{0, 1\}$ where $y_i=0$ indicates that $x_i$ is normal and $y_i=1$ indicates $x_i$ is anomalous. Using the point-estimate variant of the EM algorithm \cite{bryant1978asymptotic} \cite{celeux1992classification} \cite{gupta2011theory}, the E-step at the $m$-th iteration can be written as 

\begin{equation}\label{eq:e_step_all_data}
Y^{(m)} = \argmax_{Y} p(Y|X, \theta^{(m-1)}),
\end{equation}

\noindent where $\theta^{(m)}$ denotes the parameters of the model at the $m$-th iteration. The corresponding M-step is given by

\begin{equation}\label{eq:m_step_all_data}
\theta^{(m)} = \argmax_{\theta} p(Y^{(m)}|\theta).
\end{equation}

\noindent In detail, we choose $Y$ which maximizes the likelihood of the parameters $\theta$ given $X$ in the E-step. In the M-step, we update the parameters $\theta$ to be more inline with the new values of $Y$. 

Under the plug-and-play constraints, only one sample of the input data $x_t$ of the entire dataset $X$ is available at time $t$. We cannot wait for the whole dataset to become available, store it in a large disk and repeatedly perform expensive parameter updates over the whole dataset. Instead, the parameters are updated with every input sample $x_t$. Thus, the E-step can be written as 

\begin{equation}\label{eq:e_step_ours}
y_t = \argmax_{y \in \{0, 1\}}p(y|x_t, \theta_{t-1}),
\end{equation}

\noindent where $\theta_t$ denotes the model parameters at time $t$, and $y_t=0$ and $y_t=1$ indicate $x_t$ is normal or anomalous, respectively. The corresponding M-step then becomes

\begin{equation}\label{eq:m_step_ours}
\theta_{t} = \argmax_\theta p(y_t|\theta).
\end{equation}

As new data is continuously streaming in, we have limited time to update the parameters of the neural network as the update needs to be performed before every new sample. Hence, we do not search for the optimal set of parameters $\theta$ with every sample. Rather, we perform a single step of gradient descent in the maximization step, i.e., 

\begin{equation}\label{eq:m_step_our_time}
\begin{aligned}
\theta_t^{(D)} &= \theta_{t-1}^{(D)} - \epsilon \nabla_{\theta_{t-1}^{(D)}}L_t^{(D)} \\
\theta_t^{(G)} &= \theta_{t-1}^{(G)} - \epsilon \nabla_{\theta_{t-1}^{(G)}}L_t^{(O)},
\end{aligned}
\end{equation}

\noindent where $\epsilon$ is the learning rate to update the parameters of the neural network, $\theta^{(D)}$ denotes the parameters of the discriminator, $\theta^{(G)}$ denotes the parameters of the generator and $\theta = \theta^{(G)} \cup \theta^{(D)}$. Note that a single step of gradient descent includes both the discriminator and generator parameter updates. Performing a single step of gradient descent to approximate the optimal network parameters has been used in the past as well \cite{liu2019darts}. The parameters converge as more samples are observed by the neural network.  

\begin{figure}
\centering
\includegraphics[width=0.9\columnwidth]{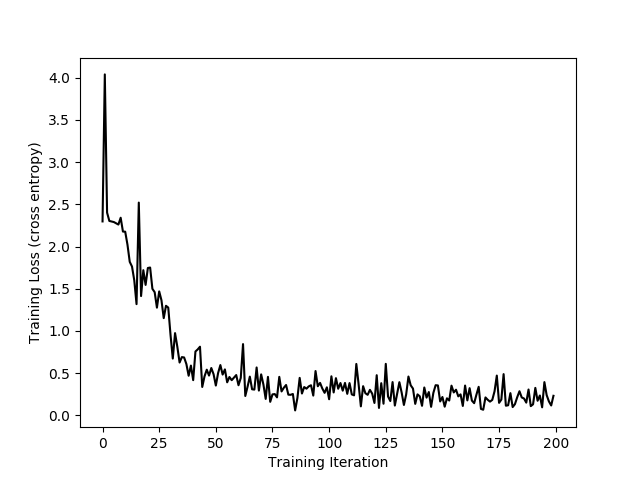}
\caption{Training loss of an MNIST digits classifier over training iterations.}
\label{fig:mnist_train_loss}
\end{figure}

For anomaly detection problems, theoretically it is not possible to learn or model the anomalous samples; otherwise, it becomes a binary classification problem where labeled training data from each class is used to train the neural network. As discussed in the introduction, anomalies can be constituted by any object or behavior that is not normal, making it impossible for a single model to represent anomalies. Therefore, we update the parameters of the model only if a sample is declared normal. More specifically, parameters are updated if $p(y=0|x_t, \theta_t)$ is greater than a certain threshold. If CAD has learned to model the normal data then the reconstruction loss can be used to detect normal and anomalous samples. In other words, the reconstruction loss $L_t^{(R)}$ decreases if the probability of a sample being normal, i.e., $p(y=0|x_t, \theta_t)$ increases. Therefore, the parameters of the model can be updated if

\begin{equation}\label{eq:loss_cond_fix}
L_t^{(R)} < \tau,
\end{equation}

\noindent where $\tau$ is a constant. However, using a constant threshold is not appropriate as a single value of $\tau$ cannot deal with the variety of situations encountered during anomaly detection. Therefore, we use a time adaptive threshold to update the parameters given as

\begin{equation}\label{eq:loss_cond}
L_t^{(R)} < \mu_{t} + \tau_t,
\end{equation}

\noindent where $\mu_t$ is an estimate of recent values of $L_t^{(R)}$ and $\tau_t$ is an adaptive threshold. It is seen from Fig. \ref{fig:mnist_train_loss} that using the condition $L_t^{(R)} < L_{t-1}^{(R)}$ is quite simplistic compared to the above and does not portray the evolution of general training loss. The training loss is not always monotonically decreasing, as seen in Fig. \ref{fig:mnist_train_loss}. Therefore, it is important to include an adaptive threshold $\tau_t$ with the criterion above. 

\begin{figure}[!bt] 
\includegraphics[width=\columnwidth]{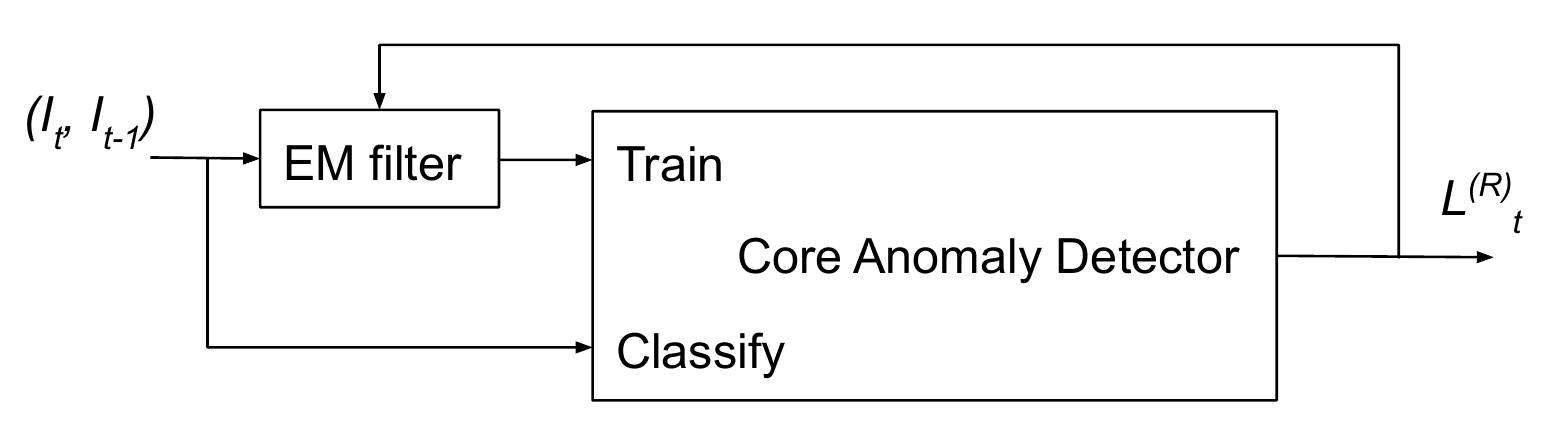}
\caption{Conceptual diagram of the overall system. The EM filter allows or prevents samples to be used for training CAD.}
\label{fig:overall_framework}
\end{figure}

Our approach to allowing samples for training CAD is quite straight-forward. If too many samples are used for training then we may have to be more strict in our choice and use a smaller value of $\tau_t$. On the other hand, if too few samples are used for updating CAD then we use a slightly larger value of $\tau_t$. Let

\begin{equation}\label{eq:delta}
\delta_t = L_t^{(R)} - \mu_t,
\end{equation}

\noindent then the mean and the threshold are updated as 

\begin{equation}\label{eq:mean_update}
\mu_{t+1} = \mu_{t} + \alpha \delta_t
\end{equation}

\noindent and 

\begin{equation}\label{eq:tol_update}
\tau_{t+1} = \tau_{t} + \alpha(\delta_t - \tau_{t}),
\end{equation}

\noindent respectively. Here $\alpha$ is the learning rate. The conceptual diagram of our approach is shown in Fig. \ref{fig:overall_framework}. All input samples are used to compute the reconstruction error $L_t^{(R)}$; however, only those which satisfy the condition in (\ref{eq:loss_cond}) are used to train CAD. Looking at the figure, one may presume that there are two forward passes, one for anomaly decision and the other for training; however, we do not perform a forward pass for training as it has already been executed, and only perform the backward pass to update the parameters of CAD using (\ref{eq:m_step_our_time}). Although we have used a batch size of one, generalization to larger batch sizes is straight-forward.

It is worthwhile to mention that our approach was initially inspired by the work in \cite{stauffer1999adaptive} for background subtraction. In \cite{stauffer1999adaptive}, authors model the background pixel intensities with a Gaussian Mixture Model (GMM). They use an online approximation of the K-means algorithm to estimate the parameters of the GMM. However, it should be noted that our work is significantly different from their approach or other EM-based approaches for parameter estimation. One of the key differences between our work and general EM-based parameter estimation approaches is that we do not make a prior assumption on the distribution of the data samples; we allow CAD to implicitly learn the distribution of the normal samples. We use the EM algorithm with a deep learning framework to perform online anomaly detection. Our method is summarized in Algorithm \ref{algo:overall}. It is seen that despite the complex nature of the problem, the proposed solution is relatively simple.

\begin{algorithm} 
\caption{Plug-and-Play Crowd Anomaly Detection}
\label{algo:overall}
\begin{algorithmic}
\STATE INPUT: $x_t = (I_t, I_{t-1}) \leftarrow$ Input frames
\STATE OUTPUT: $L_t^{(R)} \leftarrow$ Reconstruction loss
\FOR {all $t$}
	\STATE $I'_t = G(I_{t-1})$
	\STATE $L_t^{(R)} = L_t^{(i)} + L_t^{(g)}$ from (\ref{eq:int_loss}) and (\ref{eq:grad_loss})
	\IF {$L_t^{(R)} < \mu_{t} + \tau_{t}$}
		\STATE Update model: 			
		\STATE\hspace{\algorithmicindent}$\theta_t^{(D)} = \theta_{t-1}^{(D)} - \epsilon 								\nabla_{\theta_{t-1}^{(D)}}L_t^{(D)}$
		\STATE\hspace{\algorithmicindent}$\theta_t^{(G)} = \theta_{t-1}^{(G)} - \epsilon 								\nabla_{\theta_{t-1}^{(G)}}L_t^{(O)}$
		\STATE Update filter parameters:
			\STATE\hspace{\algorithmicindent} $\delta_t = L_t^{(R)} - \mu_t$  			
			\STATE\hspace{\algorithmicindent} $\mu_{t+1} = \mu_t + \alpha \delta_t$
			\STATE\hspace{\algorithmicindent} $\tau_{t+1} = \tau_{t} + \alpha(\delta_t - \tau_{t})$
	\ENDIF
\ENDFOR
\end{algorithmic}
\end{algorithm} 

\section{Experimental Results} \label{sec:exp_results}
\noindent CAD can be used in conventional settings, where we have separate training and testing stages. Therefore, we first present the performance of CAD in typical settings where it is trained with normal data only. Afterwards, we discuss the performance of the overall system constituted by CAD and the EM filter shown in Fig. \ref{fig:overall_framework}.

We have used three datasets in our experiments. The first two datasets, namely UCSD Ped1 and UCSD Ped2, are part of the UCSD dataset \cite{mahadevan2010anomaly}. These datasets have been obtained by recording a street from two different viewpoints. The normal videos include only pedestrians walking in the street, whereas anomalous videos include bicycles, vehicles, skaters etc. The UCSD Ped1 dataset has 34 training sequences and 36 testing sequences. All sequences have a frame resolution of $158 \times 238$, with every sequence containing 200 frames. The UCSD Ped2 dataset has 36 training and 12 testing sequences, having a frame resolution of $240 \times 360$ with each sequence consisting of 120 to 200 frames. The Avenue dataset \cite{Lu2019} is a recent dataset, which observes a specific location in a university campus. Normal videos include pedestrians walking on a specific pathway, whereas anomalous videos include pedestrians running, pedestrians walking on different pathways, people throwing objects etc. There are a total of 16 training and 21 testing videos, each with a frame resolution of $360 \times 640$. Other datasets have also been proposed for crowd anomaly detection, which include the UMN, Live Video (LV) \cite{leyva2017lv}, and UCF \cite{sultani2018real} datasets. The UMN dataset is quite small, the anomalies look very artificial, Area Under the Curve (AUC) values exceeding 99\% have already been achieved and the labelling is not accurate \cite{lee2015motion}. The LV dataset has generally very few normal frames for a given scene to train a neural network. The UCF dataset is quite large, however, it has been proposed for a different problem. The training data in the UCF dataset includes both the normal and anomalous frames. Also, the normal and anomalous samples are from different scenes in both train and test data. Therefore, it is not feasible to test a plug-and-play crowd anomaly detection system. 

The system used for the experiments has an Intel Core i7-3770 CPU operating at 3.4GHz and 16GB RAM. For our experiments, we used a single Nvidia Geforce 1080 Ti GPU with 12GB of memory. The system has Ubuntu 16.04. Code was developed in Python, and we used the Tensorflow \cite{abadi2016tensorflow} library for both training and testing of our model. 

For our experiments, we have used fixed learning rates of $10^{-4}$ and $10^{-5}$ for the generator and discriminator, respectively. We use a smaller learning rate for the discriminator as we do not want it to overpower the generator and result in mode collapse. We used $\lambda=0.05$ as the weighting factor in (\ref{eq:ov_gen_loss}). The learning rate for the EM filter, $\alpha$, was set to 0.1. We also set a lower limit of $5\times 10^{-5}$ on the value of $\tau_t$. 

\subsection{Conventional Anomaly Detection} 
\noindent In this section, we evaluate and compare the performance of CAD against numerous recent methods. The results of CAD have been obtained under the continual learning constraint, i.e., the network has been trained for a single epoch over the training data. The results for the Avenue, Ped1 and Ped2 datasets are shown Tables \ref{tab:avenue_comparison}, \ref{tab:ped1_comparison} and \ref{tab:ped2_comparison}, respectively. From the experimental results it is seen that CAD outperforms numerous recent and classic methods. It is interesting to note that our method provides better performance compared to approaches that use deep learning, despite the fact that CAD has been trained for a single epoch over the training data.

\begin{table}[!bt] 
\caption{Equal Error Rate (EER) and Area-Under-the-Curve (AUC) over Avenue dataset}
\label{tab:avenue_comparison}
  \centering
\begin{tabular}{lccc}
  \hline\hline
  Method & EER & AUC \\
  \hline
Adaptive \cite{sikdar2019adaptive}		& N.A. & 80.99\% \\
Conv. AE \cite{hasan2016learning}            	& 25.1\% & 70.2\%     \\
ConvLSTM-AE \cite{luo2017remembering}		& N.A. & 77.0\% \\
Discriminative \cite{del2016discriminative} 	& N.A. & 78.3 \% \\
GMM \cite{vu2017energy}				& 35.84\% & 67.27\% \\
MLAD \cite{vu2019robust}			& 36.38\% & 71.54\% \\			
OC-SVM \cite{vu2017energy}			& 33.87\% & 71.66\% \\
S-RBM \cite{vu2017energy}			& 78.76\% & 27.21\% \\
Unmasking \cite{tudor2017unmasking}		& N.A. & 80.6\% \\
CAD	                                        &28.0\% & 80.49\% \\

\hline\hline
\end{tabular}

\end{table}

\begin{table}[!bt] 
\centering
\caption{EER and AUC comparison over UCSD Ped1 dataset}
\label{tab:ped1_comparison}
  \centering
\begin{tabular}{lccc}
  \hline\hline
  Method 					& EER & AUC \\
  \hline
  Conv. AE \cite{hasan2016learning}            	&27.9\%&81.0\%    \\
ConvLSTM-AE \cite{luo2017remembering}		& N.A. & 75.5\% \\
D-IncSFA \cite{hu2016video}			& 32 \% & N.A. \\
GMM \cite{vu2017energy}				& 38.88\% & 60.33\% \\
HOFM \cite{colque2015histograms}		& 33.3\% & 71.5\% \\
HOFME \cite{colque2017histograms}               &33.1\%&72.7\%             \\
MDT \cite{mahadevan2010anomaly}                 & 25\% &25\% \\
MDT-spatial \cite{li2014anomaly}                &43.8\% & 28.7\% \\
MPPCA \cite{kim2009observe}                     &35.6\% & N.A. \\
OC-SVM \cite{vu2017energy}			& 42.97\% & 59.06\% \\
S-RBM \cite{vu2017energy}			&35.40\% & 70.25\% \\
Social Force \cite{mehran2009abnormal}          &36.5\%& N.A.   \\
Two-part Dict. \cite{masoudirad2017anomaly}	& 30\% & N.A. \\
Unmasking \cite{tudor2017unmasking}		& N.A. & 68.4\% \\

CAD	                                     	&27.1\%&79.67\%  \\

  \hline\hline
\end{tabular}
\end{table}

\begin{table}[!bt] 
\centering
\caption{EER and AUC comparison over UCSD Ped2 dataset}
\label{tab:ped2_comparison}
  \centering
\begin{tabular}{lccc}
  \hline\hline
  Method & EER & AUC \\
  \hline
AMDN \cite{xu2017detecting}			    & 17.0\% & 90.8\% \\
Binary Features \cite{leyva2017abnormal}	    & 21.2\% & N.A. \\
Compact Feature \cite{leyva2017video}		    & 19.2\% & N.A. \\
Conv. AE \cite{hasan2016learning}            	    & 21.7\% & 90\%     \\
ConvLSTM-AE \cite{luo2017remembering}		    & N.A. & 88.1\% \\
DOC \cite{sun2017abnormal}			    & 16.1\% & 91.1\% \\
Fast CNN \cite{SABOKROU2018}                        & 11\% & N.A.\\
GMM  \cite{rojas2016abnormal}		            & N.A. & 90.3\% \\
H-MDT CRF \cite{li2014anomaly}                      & 18.5\% & N.A.   \\
HOFME \cite{colque2017histograms}                   & 20\% &   87.5\%           \\
HOFM \cite{colque2015histograms}		    & 19.0\% & 19.0\% \\
HOT-FS \cite{mousavi2015analyzing}		    & 21.20\% & N.A. \\
iHOT: FS \cite{mousavi2015abnormality}		    & 16.5\% & N.A. \\
Inc. Coding \cite{dutta2015online}		    & 22.3\% & N.A. \\
MDT \cite{mahadevan2010anomaly}                     & 25\% & N.A. \\
MDT-spatial \cite{li2014anomaly}                    & 28.7\% & N.A. \\
MPPCA \cite{kim2009observe}                         & 35.8\% & N.A.\\
RBM \cite{vu2017energy}				    & 16.47\% & 86.43\% \\
One Class SVM 	\cite{lin2015anomaly}		    & 20\% & N.A. \\
Social Force \cite{mehran2009abnormal}              & 35.0\% & N.A.\\
Social Net \cite{chaker2017social}		    & N.A. & 87.9\% \\
Sparse AE \cite{sabokrou2016video}		    & 15\% & N.A. \\
Sparse Dict. (mini batch) \cite{Lu2019}		    & N.A.  &  87\% \\
SpatioTemporal Context \cite{li2015spatio}          & 20\% & 89.1\% \\
SpatioTemporal AE \cite{chong2017abnormal}	    & 12.0\% & 87.4 \% \\
Stacked RNN \cite{luo2017revisit}		    & N.A. & 92.21\% \\
Struct. Anal. \cite{yuan2014online}		    & N.A. & 92.5\% \\
TCP \cite{ravanbakhsh2016plug}			    & 18\% & 88.4\% \\
Unmasking \cite{tudor2017unmasking}		    & N.A. & 82.2\% \\ 	

CAD	                                            &14.5\% & 92.84\% \\

\hline\hline
\end{tabular}
\end{table}

\begin{figure*}[!bt]
\centering
\subfloat[]{\includegraphics[width=0.22\textwidth]{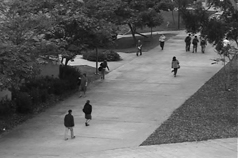}}\hfil
\subfloat[]{\includegraphics[width=0.22\textwidth]{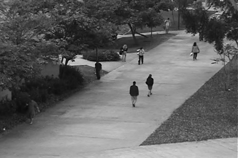}}\hfil
\subfloat[]{\includegraphics[width=0.22\textwidth]{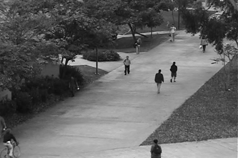}}\hfil
\subfloat[]{\includegraphics[width=0.22\textwidth]{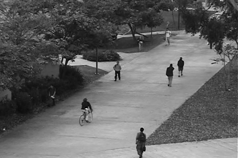}}\hfil
\subfloat[]{\includegraphics[width=0.95\textwidth]{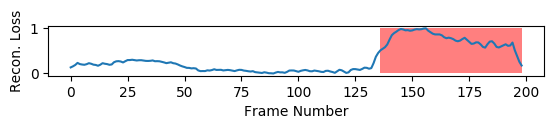}}\hfil
\caption{Frames (a) 25, (b) 100, (c) 140 and (d) 160 of the 15\textsuperscript{th} test video of Ped1 dataset. (e) Reconstruction loss for the test video with CAD. Shaded regions indicate anomalous frames.}
\label{fig:ped1_vid_result}
\end{figure*}

\begin{figure*}[!bt]
\centering
\subfloat[]{\includegraphics[width=0.22\textwidth]{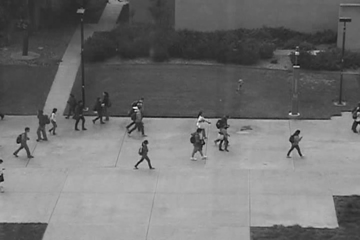}}\hfil
\subfloat[]{\includegraphics[width=0.22\textwidth]{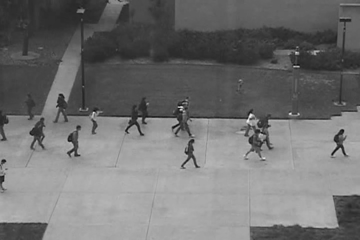}}\hfil
\subfloat[]{\includegraphics[width=0.22\textwidth]{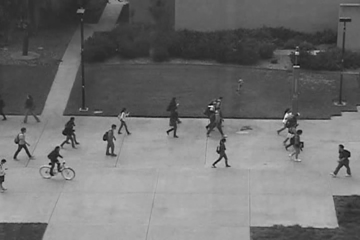}}\hfil
\subfloat[]{\includegraphics[width=0.22\textwidth]{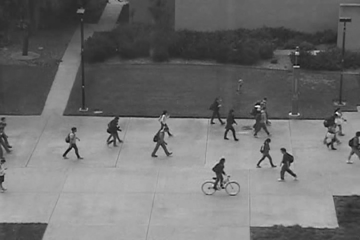}}\hfil
\subfloat[]{\includegraphics[width=0.95\textwidth]{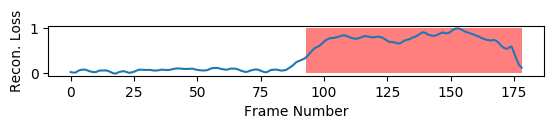}}\hfil
\caption{Frames (a) 25, (b) 75, (c) 110 and (d) 160 of the 2\textsuperscript{nd} test video of Ped2 dataset. (e) Reconstruction loss for the test video with CAD. Shaded regions indicate anomalous frames.}
\label{fig:ped2_vid_result}
\end{figure*}

\begin{figure*}[!bt]
\centering
\subfloat[]{\includegraphics[width=0.22\textwidth]{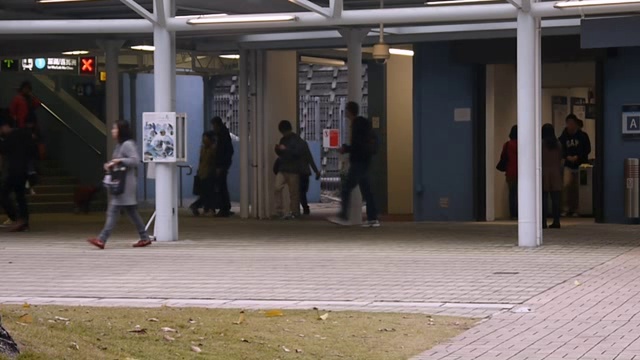}}\hfil
\subfloat[]{\includegraphics[width=0.22\textwidth]{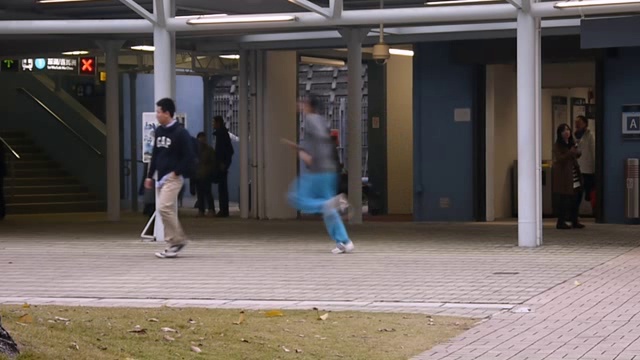}}\hfil
\subfloat[]{\includegraphics[width=0.22\textwidth]{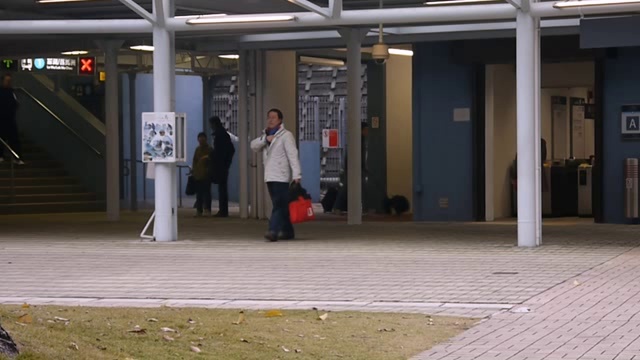}}\hfil
\subfloat[]{\includegraphics[width=0.22\textwidth]{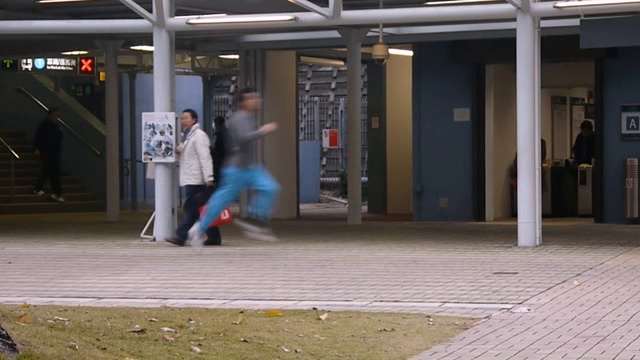}}\hfil
\subfloat[]{\includegraphics[width=0.95\textwidth]{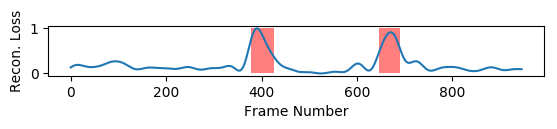}}\hfil
\caption{Frames (a) 100, (b) 400, (c) 600 and (d) 665 of the 4\textsuperscript{th} test video of Avenue dataset. (e) Reconstruction loss for the test video with CAD. Shaded regions indicate anomalous frames.}
\label{fig:avenue_vid_result}
\end{figure*}

\begin{figure}[!bt]
\includegraphics[width=0.9\columnwidth]{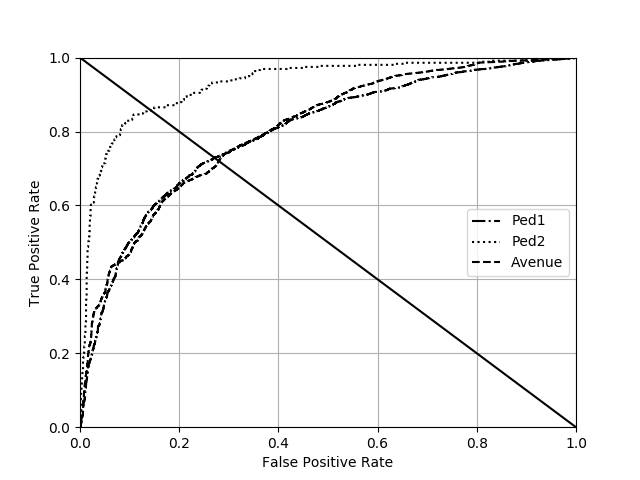}
\caption{Reciever Operating Characteristics (ROC) of CAD with different datasets.}
\label{fig:roc_curve}
\end{figure}

Some visual examples of the performance of CAD are shown in Fig. \ref{fig:ped1_vid_result}, \ref{fig:ped2_vid_result} and \ref{fig:avenue_vid_result}. It is observed that the reconstruction loss is higher at anomalous frames compared to the normal frames for these examples. The ROC curves are also shown in Fig. \ref{fig:roc_curve}. The figure shows that the proposed method shows competitive performance on the Avenue and Ped1 datasets, however, its performance on the Ped2 dataset is significantly better.

We also compare the performance of CAD with other methods when training is performed for a single epoch only. The results are shown in Table \ref{tab:one_epoch_perf}. We used a re-implementation of \cite{ionescu2019object} provided at \cite{object_centrci_VAD} in these experiments. It is seen that our method outperforms these methods with a single epoch of training, except for \cite{liu2018future} on the Avenue dataset. These results show the feasibility of the proposed CAD for continual learning.

\begin{table}[!bt]
\centering
\caption{AUC Comparison with Single Epoch Training}
\label{tab:one_epoch_perf}
\centering
\begin{tabular}{lcccc}
\hline \hline
Method 					& UCSD Ped1 & UCSD Ped2 & Avenue \\
\hline 
Conv LSTM-AE \cite{luo2017remembering} 	& 75.02\% & 80.58\% & 77.07\% \\
Frame Prediction \cite{liu2018future} 	& 75.29\% & 92.1\% & 83.32\% \\
Obj. Centric \cite{ionescu2019object}	& 46.73\% & 67.16\% & 53.66\% \\
Stacked RNN \cite{luo2017revisit} 	& 61.66\% & 72.97\% & 79.76\% \\

CAD	 				& 79.67\% & 92.84\% & 80.49\% \\
\hline \hline
\end{tabular}
\end{table}

For further insights, we analyze the performance of CAD with different categories of anomalies in different datasets. In Table \ref{tab:cat_analysis}, we provide the average AUC of test sequences containing a specific form of anomalous behavior. From the table, it is seen that CAD performs well when an anomaly is due to change in motion behavior. For other categories, there is room for improvement for the future.

\begin{table}[!bt]
\centering
\caption{Category-wise performance of CAD}
\label{tab:cat_analysis}
\begin{tabular}{l|l|c}
\hline\hline
Dataset				& Anomalous Activity  	& AUC 		\\ \hline
\multirow{7}{*}{UCSD Ped1}	& Cycling		& 88.81\%	\\
				& Running		& 99.74\%	\\
				& Skateboarding	& 79.91\%	\\
				& Standing		& 61.85\%	\\
				& Vehicles		& 93.13\%	\\
				& Wheelchair		& 41.31\%	\\
				& Wrong direction	& 74.18\%	\\ \hline
\multirow{3}{*}{UCSD Ped2}	& Cycling		& 99.46\%	\\
				& Skateboarding	& 57.71\%	\\
				& Vehicles		& 100\%		\\ \hline
\multirow{6}{*}{Avenue}		& Dance moves		& 52.61\%	\\
				& Jumping kid		& 43.2\%	\\
				& Running		& 85.36\%	\\
				& Throwing a bag	& 78.69\%	\\
				& Throwing papers	& 75.55\%	\\
				& Wrong direction	& 71.2\%	\\
\hline \hline
\end{tabular}
\end{table}

\subsection{Plug-And-Play Crowd Anomaly Detection}
\noindent After evaluating the performance of CAD, in this section we evaluate the performance of the overall framework composed of CAD and the EM filter. For these experiments, we mixed the normal and anomalous frames at a controlled rate to set up the experimental environment. For each dataset, we first divide the whole dataset (training and test videos) into clips of normal $X_N$ and anomalous $X_A$ sequences. To generate the input data stream for the overall framework of Fig. \ref{fig:overall_framework}, we use a random selector. The selector generates random numbers $r_t\in[0,1]$ at time $t$. If the required portion of anomaly in the stream is $s$ then the input $x_t$ is obtained as 

\begin{equation} \label{eq:ano_mix}
\begin{split}
x_t &\in X_A - X_A^{(p)}~~~~~~\textrm{if}~~~~~~r_t<s \\
x_t &\in X_N - X_N^{(p)}~~~~~~\textrm{otherwise},
\end{split}
\end{equation}

\noindent where $X_A^{(p)}$ and $X_N^{(p)}$ are the sets of anomalous and normal samples, respectively, previously added to the video stream as follows.

\begin{equation} \label{eq:ano_mix_set}
\begin{split}
X_A^{(p)}&:= X_A^{(p)} \cup \{x_t\}~~~~~~\textrm{if}~~~~~~r_t<s \\
X_N^{(p)}&:= X_N^{(p)} \cup \{x_t\}~~~~~~\textrm{otherwise}.
\end{split}
\end{equation}

\noindent In detail, the samples passed to the video stream are removed from the dataset so that the video stream does not contain redundant samples. The total number of samples used in the experiments is equal to the total number of samples in the training data of a given dataset. 

\begin{figure}[!bt]
\centering
\subfloat[]{\includegraphics[width=0.5\columnwidth]{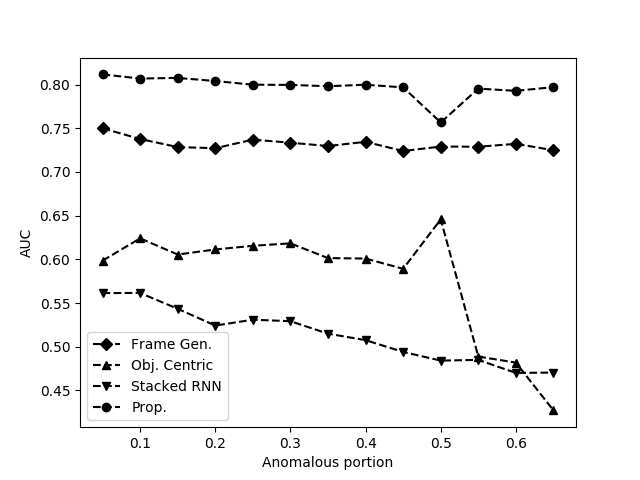}}\hfil
\subfloat[]{\includegraphics[width=0.5\columnwidth]{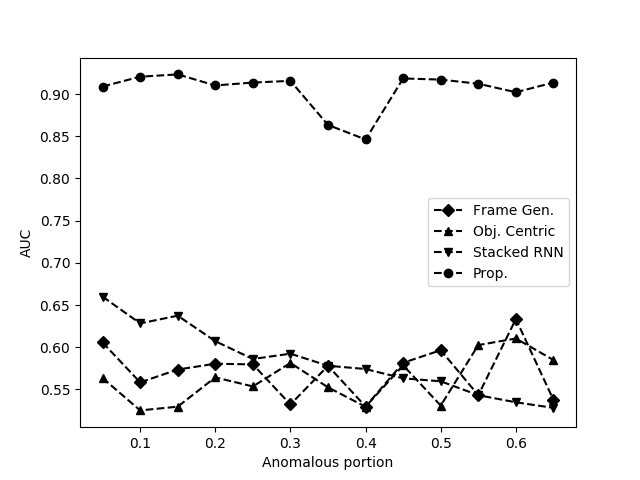}}\hfil
\subfloat[]{\includegraphics[width=0.5\columnwidth]{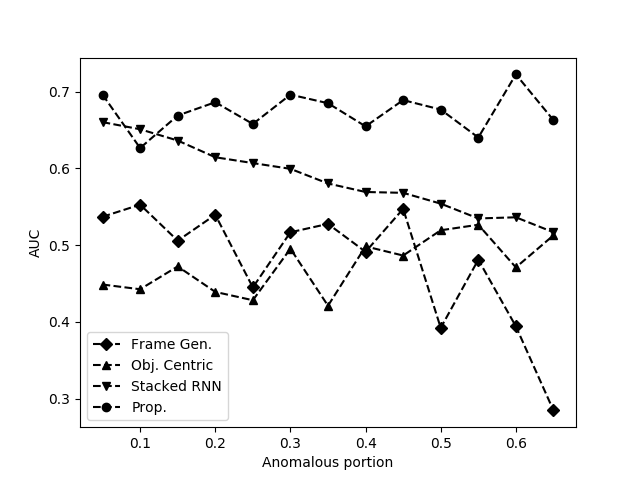}}\hfil
\caption{Anomaly detection performance with mixed data streams from the (a) UCSD Ped1, (b) UCSD Ped2, and (c) Avenue datasets.}
\label{fig:real_time_auc}
\end{figure}

The AUC for different settings of anomaly are shown in Fig. \ref{fig:real_time_auc}. The results  are shown after conducting the experiments thrice and averaging the results. For comparison, we also show the results of Future Frame Generation \cite{liu2018future}, Object Centric \cite{ionescu2019object} and Stacked RNN \cite{luo2017revisit} methods . From the results, it is seen that the performance of other methods is lower compared to our method, and generally degrades with the anomalous portion of data in the input stream. On the other hand, the performance of our method is generally better and remains consistent despite changes in the anomalous portion of the input stream. The average AUC of our method is compared against other methods in Table \ref{tab:avg_pract_auc}. The results show that our method outperforms these methods under the plug-and-play constraints for crowd anomaly detection. However, there is room for improvement in the results on the Avenue dataset. Note that the proposed approach provides 10fps in our experimental setup.

One interesting question that arises as a result of these experiments is that how can the overall anomaly detection system work if the anomalous samples exceed the normal samples. In this scenario, we expect CAD to treat the anomalous samples as normal as they are in a majority. Our intuition is that although the portion of anomalous samples in the input stream exceeds the normal samples, the anomalous samples come from different sources. The anomalous samples are composed of running pedestrians, bicycles, vehicles, skaters, and people walking on different pathways whereas the normal samples only include pedestrians walking on a particular pathway. The number of normal samples is greater than the number of samples of each individual category of anomaly. Therefore, the system is capable of rejecting individual instances of anomalous frames from training the neural network as these anomalous samples by each category are still fewer than the normal samples. Technically, the individual instances of anomaly cannot exceed the normal samples, otherwise, the definition of normal samples should be changed.

\begin{table}[!bt]
\centering
\caption{AUC with Plug-and-Play Anomaly Detection}
\label{tab:avg_pract_auc}
\centering
\begin{tabular}{lcccc}
\hline \hline
Method 					& UCSD Ped1 & UCSD Ped2 & Avenue \\
\hline
Frame Prediction \cite{liu2018future}	& 73.2\% & 57.1\% & 47.81\% \\
Obj. Centric \cite{ionescu2019object} 	& 57.75\% & 56.20\% & 47.39\% \\
Stacked RNN \cite{luo2017revisit} 	& 51.36\% & 58.40\% & 58.68\% \\
Proposed				& 79.24\% & 90.43\% & 67.41\% \\
\hline \hline
\end{tabular}
\end{table}

\begin{figure*}[!bt]
\centering
\subfloat[]{\includegraphics[width=0.95\textwidth, height=1.5in]{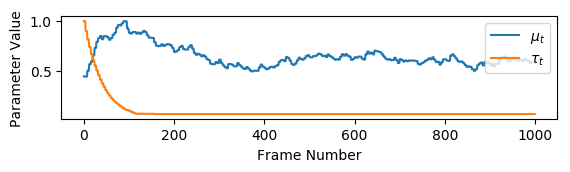}}\hfil
\subfloat[]{\includegraphics[width=0.95\textwidth, height=1.5in]{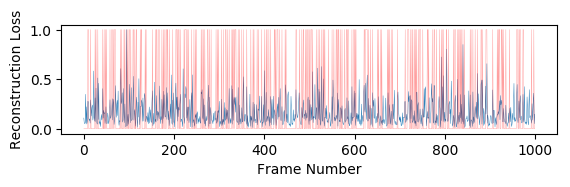}}\hfil
\caption{Values of the (a) parameters of the EM filter and (b) reconstruction loss over time with the Ped1 dataset (red plot indicates anomalous inputs). Parameters remain relatively stable after an initial transient despite changes in reconstruction loss.}
\label{fig:ped1_param_evo}
\end{figure*}

\begin{figure*}[!bt]
\centering
\subfloat[]{\includegraphics[width=0.95\textwidth, height=1.5in]{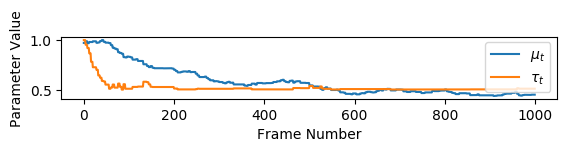}}\hfil
\subfloat[]{\includegraphics[width=0.95\textwidth, height=1.5in]{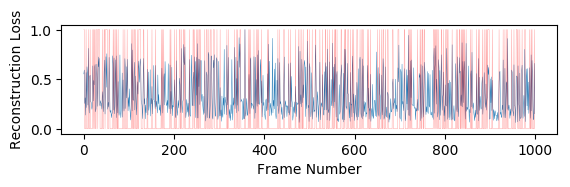}}\hfil
\caption{Values of the (a) parameters of the EM filter and (b) reconstruction loss over time with the Ped2 dataset (red plot indicates anomalous inputs). Parameters remain relatively stable after an initial transient despite changes in reconstruction loss.}
\label{fig:ped2_param_evo}
\end{figure*}

To observe the behavior of the parameters $\mu_t$ and $\tau_t$, we have plotted them in Fig. \ref{fig:ped1_param_evo} and \ref{fig:ped2_param_evo} for the Ped1 and Ped2 datasets, respectively. We have also plotted the corresponding loss values. These results are for an anomalous portion of 25\% in the input stream. Similar results are observed for other values as well. The parameter values are max-normalized for better viewing. From the figures, it is seen that after an initial transient period, the parameters settle down to relatively stable values despite the large variation in the reconstruction loss $L_t^{(R)}$. 

\subsection{Effect of the EM Filter}
\begin{figure*}
\centering
\includegraphics[width=0.9\textwidth]{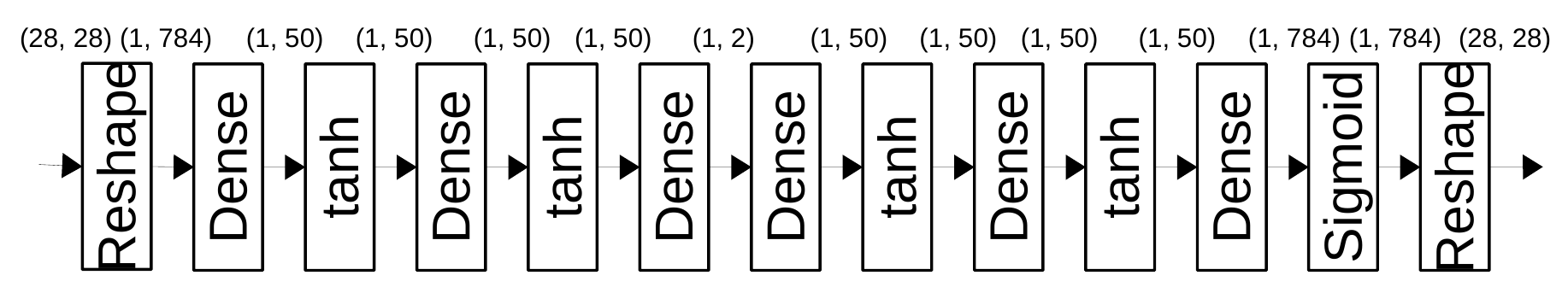}
\caption{The autoencoder used for experiments with MNIST digits.}
\label{fig:mnist_ae}
\end{figure*}

\noindent In this subsection, we further analyze the effects of the EM filter. These experiments have been performed on the MNIST dataset \cite{lecun1998gradient}. In our experiments, we treated the images corresponding to digits 0 and 1 as normal and anomalous, respectively.  Normal and anomalous data were mixed and input to the autoencoder as described in the previous subsection. Note that the labels of the digits were used for evaluation only. We used a simple autoencoder, shown in Fig. \ref{fig:mnist_ae}, with the EM filter. The reconstruction loss of the input image was used for both training and anomaly detection. Learning rates of $\epsilon=10^{-4}$ and $\alpha=0.25$ were used to update the autoencoder and the EM filter, respectively. Training was performed for a single epoch.

\begin{figure}
\centering
\includegraphics[width=0.9\columnwidth]{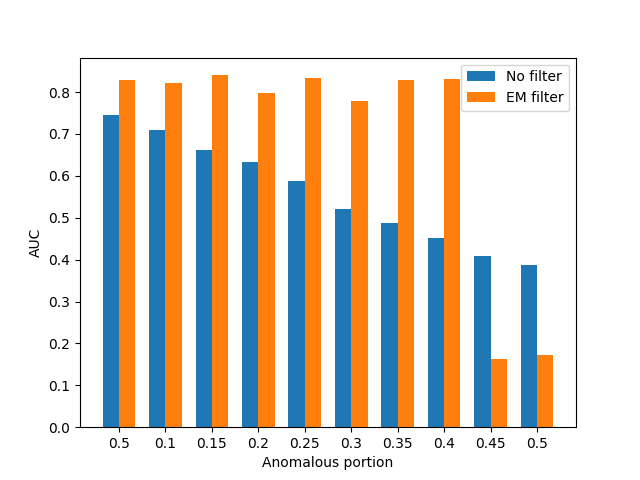}
\caption{Comparison of the AUC values with and without the EM filter on the MNIST dataset.}
\label{fig:mnist_auc_comp}
\end{figure}

The AUC values with and without the proposed EM filter are shown in Fig. \ref{fig:mnist_auc_comp}. It is seen from the figure that the performance of the autoencoder without the EM filter degrades as the anomalous portion in the input stream is increased. On the contrary, the performance of the autoencoder is generally better and consistent at different portions of anomaly in the input stream. However, the AUC with the EM filter decreases steeply when the input stream has around 45\% anomalous samples.

\begin{figure}
\centering
\subfloat[]{\includegraphics[width=0.9\columnwidth]{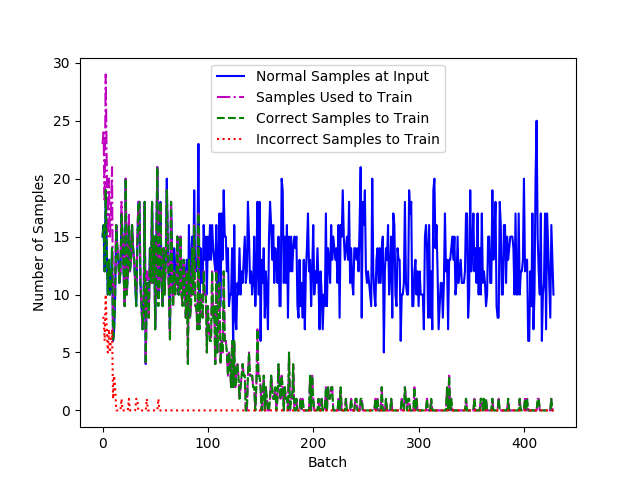}}\hfil
\subfloat[]{\includegraphics[width=0.9\columnwidth]{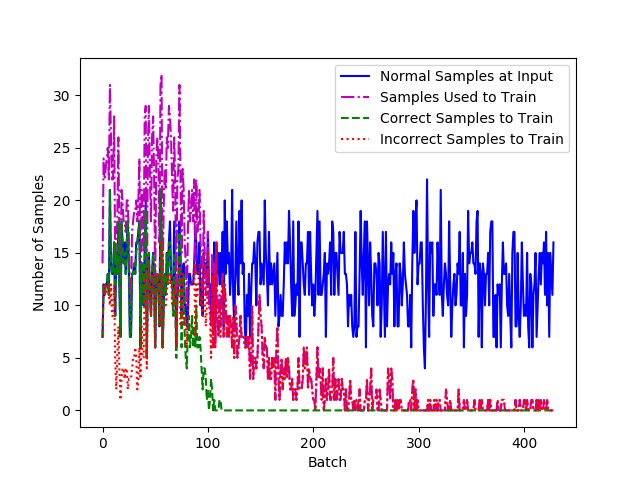}}
\caption{Analysis of the EM filter with anomalous portions of (a) 25\% and (b) 50\% in the input data. \emph{Normal Samples at Input} indicates the number of normal samples at the input of the EM filter, \emph{Samples Used to Train} denotes the number of samples allowed by the EM filter to train the autoencoder, \emph{Correct Samples to Train} and \emph{Incorrect samples to Train} are the samples correctly and incorrectly chosen to train the autoencoder, respectively.} 
\label{fig:mnist_sample_analysis}
\end{figure}

For further insights, we graphically show how accurately the EM filter can filter-out anomalous samples from training the neural network. The results are shown in Fig. \ref{fig:mnist_sample_analysis}. The figure shows two results; first with an anomalous portion of 25\% when the EM filters performs well and second with an anomalous portion of 50\% when the EM filter performs poorly. It is seen that at an anomalous portion of 25\% in the input stream, the EM filter initially chooses some anomalous samples to train the autoencoder. However, the number of anomalous samples per batch used for training is reduced to zero with the passage of time and all the samples used to train the autoencoder are normal samples. The performance at an anomalous portion of 50\% in the input stream is almost the opposite. For the initial few batches, both normal and anomalous samples are allowed by the EM filter to train the autoencoder. However, after a few batches, the EM filter only allows the anomalous samples to train the autoencoder while rejecting all the normal samples. This is because the number of anomalous and normal samples in the input stream are roughly the same, and the anomalous samples are quite similar as these are all images of the digit 1. There is an equal chance that the EM filter either chooses images of digit 0 or images of digit 1 to train the autoencoder. Interestingly, the autoencoder is still capable of distinguishing the normal and anomalous samples. In detail, the AUC with 50\% anomalous samples in the input stream is 17.18\%, which is even lower than not using any filter at all at the input. However, if a smaller reconstruction loss is used to indicate an anomalous sample and a larger reconstruction loss is used to indicate a normal sample, then the AUC value is increased to 82.81\%. 

Despite numerous advantages, there is still room for improvement. Fig. \ref{fig:mnist_sample_analysis} show that the EM filter rejects many normal samples after an initial period. Samples used for training the autoencoder are generally fewer than the normal samples at the input of the EM filter. As future work, the training sample efficiency of the EM filter can be increased to further enhance the performance.

\section{Conclusion}
\noindent In this paper, we aim to develop methods for a plug-and-play crowd anomaly detection system. The system should start learning from scratch about normal and anomalous events in a given scene as soon as it is deployed. This places numerous constraints on such a system, which are much different from conventional anomaly detection frameworks. The proposed method shows competitive performance when trained with a single epoch on the training data. In situations where the anomalous and normal data are not separated and observed as a single stream, the proposed method shows superior performance for crowd anomaly detection compared to other deep learning-based methods. To our knowledge, this is the first work that proposes an online crowd anomaly detection system with deep learning.

\bibliographystyle{IEEEtran}
\bibliography{IEEEabrv,bibItemsAll}

\end{document}